\documentclass[pdflatex,sn-mathphys]{sn-jnl}

\jyear{2021}%

\theoremstyle{thmstyleone}%
\theoremstyle{thmstyletwo}%

\theoremstyle{thmstylethree}%

\usepackage{caption}
\usepackage{subcaption}
\usepackage{amsmath,amssymb}
\usepackage[columnwise]{lineno}
\usepackage{textcmds}

\usepackage{afterpage}
\usepackage{algorithm}
\usepackage{algpseudocode}
\usepackage{color}

\begin{document}

\title[Severe Damage Recovery in Evolving Soft Robots]{Severe Damage Recovery in Evolving Soft Robots through  Differentiable Programming}


\author*[1]{\fnm{Kazuya} \sur{Horibe}}\email{horibe@irl.sys.es.osaka-u.ac.jp}

\author[2]{\fnm{Kathryn} \sur{Walker}}\email{kwal@itu.dk}
\equalcont{These authors contributed equally to this work.}

\author[2]{\fnm{Rasmus} \sur{Berg Palm}}\email{rasmb@itu.dk}
\equalcont{These authors contributed equally to this work.}

\author[2]{\fnm{Shyam \sur{Sudhakaran}}}\email{shyamsnair@protonmail.com}
\equalcont{These authors contributed equally to this work.}

\author[2]{\fnm{Sebastian} \sur{Risi}}\email{sebr@itu.dk}

\affil*[1]{\orgdiv{Department of Systems Innovation}, \orgname{Osaka University}, \orgaddress{\street{1-3 Machikaneyama}, \city{Toyonaka}, \postcode{560-8531}, \state{Osaka}, \country{Japan}}}

\affil[2]{\orgdiv{Department of Digital Design}, \orgname{IT University of Copenhagen}, \orgaddress{\street{Rued Langgaards Vej 7}, \city{Copenhagen}, \postcode{DK 2300},  \country{Denmark}}}


\abstract{Biological systems are very robust to morphological damage, but artificial systems (robots) are currently not. In this paper we present a system based on neural cellular automata, in which locomoting robots are evolved and then given the ability to regenerate their morphology from damage through gradient-based training. Our approach thus combines the benefits of evolution to discover a wide range of different robot morphologies, with the efficiency of supervised training for robustness through differentiable update rules. The resulting neural cellular automata are able to grow virtual robots capable of regaining more than 80\% of their functionality, even after severe types of morphological damage.}




\maketitle
\section{Introduction}\label{sec1}
Within the natural world, evolution has created a diverse range of robust, adaptive organisms that are able to survive damage. While this adaptation is sometimes due to a change in behaviour (or control system), in many cases these organisms rely on morphological regeneration. In these instances, the organisms repair and/or reconfigure their morphology in response to damage or changes in components \cite{carlson2011principles}. 

The amount of regeneration for which an organism is capable varies from species to species. For many years, gardeners have deliberately damaged (pruned) their plants to encourage plant re-growth in a particular direction or increase fruit yield \cite{wade2009basic,davis1993spacing}. Plant propagation through cutting (where a completely new plant is grown from part of an existing plants stem or root) is also common practise \cite{hartmann1975plant}. Whilst pruning can be considered a chore by many \cite{wade2009basic}, it does help to highlight the power of nature's morphological regeneration.   

Morphological regeneration does not only occur in plants but also in animals. For instance, salamanders are capable of regenerating an amputated leg \cite{vieira2020advancements}. The simple organisms Hydra and Planaria are capable of complete morphological repair, regardless of which body part is removed \cite{levin2019endogenous,vogg2019model}. Even within humans, the liver organ is able to regenerate and replace lost liver tissue via growth from the remaining cells \cite{fausto2006liver}.

In contrast to biological organisms, artificial robotic systems are fragile and even a small amount of damage can severely impact their performance. Furthermore, despite its commonplace within nature, the majority of examples of robots adapting to damage focus on a change in control system, not morphology. This potentially limits their robustness compared with the wide range of damage natural systems are able to recover from.

\begin{figure}[htpb!]
    \centering
    \includegraphics[width=1\linewidth]{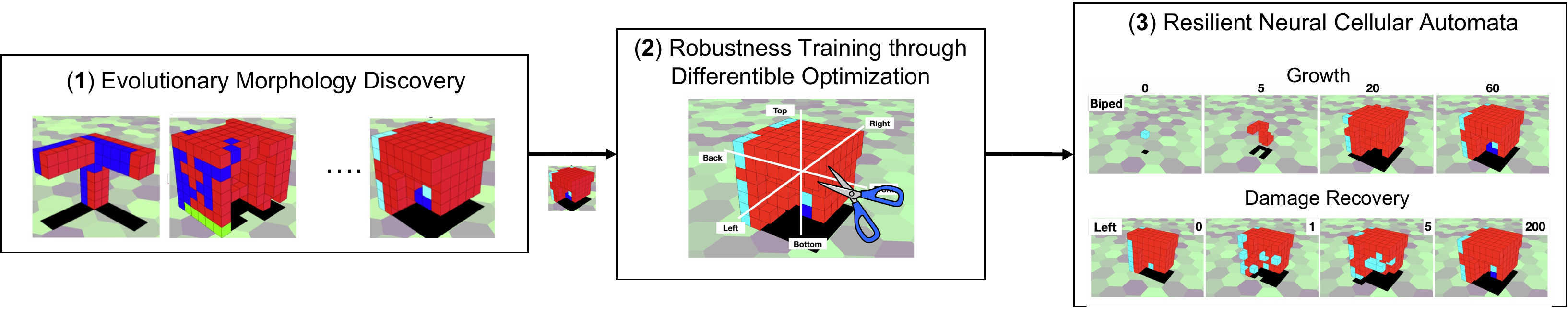}
    \caption{Approach Overview. (1) A diversity of morphologies are discovered through evolutionary optimization. (2) A neural cellular automata is trained to regrow a target morphology found by evolution under different damages. (3) The resulting NCA is able to grow a soft robot, while being able to recover from extreme forms of damage. }
    \label{fig:overview}
\end{figure}

While the mechanisms behind natural morphological regeneration are still not fully understood, artificial methods such as neural cellular automata have proven successful tools for their modelling. 
For instance, the work presented in this paper is an extension of our previous conference paper \cite{horibe2021regenerating}, where we evolve neural cellular automata to grow locomoting soft robots capable of some morphological regeneration. 
Neural cellular automata are based on the simpler cellular automata. These consist of a regular grid of cells where each cell can be in any one of a finite set of states, cell states are then updated based on information from their neighbour's states and simple rule sets. In neural cellular automata, these simple rules are replaced by artificial neural networks.

In this paper, we extend our initial investigation and use a more efficient way of training neural cellular automata for robustness, which is based on gradient-based optimization. An overview of the approach is shown in Figure~\ref{fig:overview}. As before \cite{horibe2021regenerating}, we first use an evolutionary algorithm to discover neural cellular automata that can grow the control and morphology of a diversity of soft voxel-based robots. The NCA for a particular robot is then efficiently trained through differentiable programming (i.e.\ gradient-based optimization) to grow a particular robot while being able to recover from various forms of damage.

In contrast to previous work on training for regeneration through gradient-based optimization  \cite{sudhakaran2021growing,mordvintsev2020growing}, here the target artefacts are evolved instead of being hand-designed. This approach thus combines the creativity of artificial evolution with the efficiency of supervised training, opening up new application in the field of morphological computation.


\subsection{Related Work}\label{related}

\subsubsection{Evolving Virtual Creatures}
In this paper we explore growing virtual soft robots, capable of regeneration, via the use of trained neural cellular automata. In the first instance we use evolutionary algorithms to grow simple soft robots, using distance travelled as our fitness function. However, the concept of using artificial evolution to design the control and morphology of virtual robots has existed for almost three decades. In his seminal work, Karl Sims evolved the body plans of simple block based robots, which interacted with their virtual environment \cite{sims1994evolving,sims1994evolvingbook}. This work was then followed by many researchers also investigating using evolutionary algorithms to design the morphology and control of virtual robots, e.g., ~\cite{dellaert1994co,eggenberger1997evolving,ostergaard2003evolving,lipson2000automatic,risi2013ribosomal}.

A common theme amongst researchers investigating evolving robot morphologies is the use of compositional pattern producing networks (CPPNs) \cite{stanley2007compositional} to describe body shape and properties  \cite{cheney2014unshackling,cheney2015evolving,cheney2018scalable,auerbach2014environmental,auerbach2010evolving}. CPPNs are a special type of artificial neural network, applied over the whole input space, allowing patterns or shapes to be produced. 

However, these previous examples only consider static morphologies, that is, they do not allow for shape changes for growth or damage recovery. 
Therefore,  researchers have more recently begun to investigate different approaches to address these limitations, which we review next. 

\subsubsection{Damage recovery/ shape change in robots}

As  mentioned previously, when trying to adapt to new environments/tasks or recover from damage, researchers have mostly explored control-based approaches instead of changing the robot's morphology  \cite{urzelai2000evolutionary,nolfi1999learning, chatzilygeroudis2018reset,cully2015robots,kano2017brittle,najarro2020meta}. The investigation into metamorphosis/morphological regeneration as a damage recovery strategy is less common, despite its potential and prevalence in nature.

Recent examples of robots adapting their morphology  to different environments, rather than control, include the work by Kriegman et al. \cite{kriegman2018interoceptive} where individual parts of the robot change stiffness in response to external stresses. Walker et al.~\cite{walker2021evolution} explore removing individual parts of the robot to sculpt morphologies based on environmental interaction.
Shah et al. published work on a physical soft robot capable of radial morphological shape change for locomotion in different environments \cite{shah2021soft}. We refer the interested reader to the extensive review paper on shape changing robots by Shah et al.~\cite{shah2021shape}.

These previous examples all refer to morphology change for environmental adaptation. However, there are also examples where morphological change has been exploited  for damage recovery. These works include that by Kriegman et al.~\cite{kriegman2019automated}, where silicone based physical voxel robots were able to recover from voxel removal. Furthermore, Xenobots, synthetic creatures designed from biological tissue \cite{kriegman2020scalable}, have shown to be capable of morphological reattachment (i.e.\  healing after insult).

\subsubsection{Neural cellular automata for growth and recovery}

Cellular automata (CA) were first proposed by Neumann and Ulam in the 1940s and consist of a regular grid of cells where each cell can be in any one of a finite set of states \cite{neumann1966theory}. Each cell determines its next state based on local information (i.e.\ the states of its neighboring cells) according to pre-defined rules. Instead of hand-designed rules, these rules can also be learned.  For example,   Miller showed that automatic recovery of simple damaged target patterns is possible by learning CA rules through genetic programming \cite{miller2004evolving}. 

CA rules can also be learned by neural networks, resulting in what is now called a neural cellular automata (NCA) \cite{wulff1992learning}.  
NCAs have been used to great effect for morphological growth and recovery in simulation. For example,  Mordvintsev et al.  trained a neural CA to grow complex two-dimensional images starting from a few initial cells through gradient-based optimization (i.e.\ differentiable programming) \cite{mordvintsev2020growing}. In addition, the authors have also successfully trained the system to reconstruct the pattern after damage (i.e.\ it was able to regrow the target pattern). The neural network in their work is a convolutional network, which lends itself to represent neural CAs \cite{gilpin2019cellular}. 

Based on Mordvintsev's work, Sudhakaran et al. \cite{sudhakaran2021growing} have trained neural cellular automata capable of growing particular target patterns in three dimensions. As with Mordvintsev's work, these 3D shapes/structures are able to regrow after significant damage and are also trained through gradient-based optimization. Both approaches \cite{mordvintsev2020growing, sudhakaran2021growing} relied on user-specific target patterns, while these target patterns are themselves evolved in our approach and then made robust against damage.

\section{Method}\label{sec2}
\subsection{Soft Robot Simulator}

For our investigations (both 2D and 3D) we use the soft robot simulator ``Voxelyze" \cite{hiller2009multi}.  In Voxelyze, a cell takes the form of a 3D cube called a voxel; adjacent voxels (or cells) are connected together by simulated beams. Voxelyze creates actuation within these soft robots by employing a sinusoidally varying global control signal (termed temperature by the Voxelyze software). Active cells expand and contract in phase with this sinusoidal temperature variation, whilst passive cells remain a constant size. The software VoxCad can then be used to visualise this. 

Our aim is to evolve soft robots capable of successful locomotion, as well as regeneration, both in 2D and 3D. Thus the fitness of a robot (in terms of locomotion) is how far the robot travelled in a fixed time period. This time period is $0.25$ seconds, or $10$ actuation cycles in 2D and for $0.5$ seconds, or $20$ actuation cycles in 3D. The evaluation times try to strike a balance between reducing computational costs while still giving sufficient time to observe interesting locomotion behaviours. Fitness is determined as the distance the robot’s center of mass moves in $0.25$ or $0.5$ seconds. The distance (or fitness) of a robot is measured in units that correspond to the length of a voxel with volume one. It should also be noted that creatures with zero voxels after their growth are automatically assigned a fitness of 0.0.

In all our experiments, the neural cellular automata is able to utilise four types of cell/voxel. The first is an active cell, denoted by the colour red, and has an expansion and contraction cycle in phase with the global control signal. We also refer to this type of cell as muscle. The second type of cell can also be thought of as muscle. Denoted by the colour green, it also expands and contracts, but this time at counter phase to the red muscle cells. Dark blue cells are passive, they do not change size. We refer to these as bone. The final cell type is denoted by a light blue colour. These are also passive but have a lower stiffness than dark blue bone voxels. 

Our choice of cell types is consistent with previous literature (Cheney et al.~\cite{cheney2014unshackling}), as is our choice of physical and environmental Voxelyze parameters. 

\begin{figure}[htpb!]
    \centering
        \begin{subfigure}[t]{1.0\linewidth}
            \includegraphics[width=\textwidth]{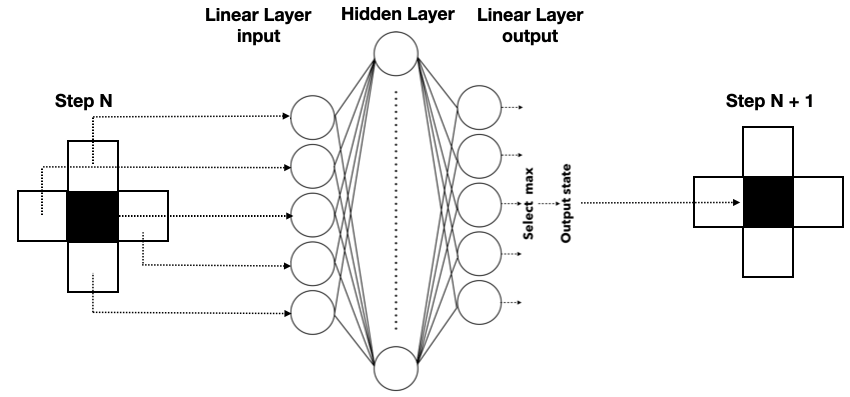}
            \caption{Evolutionary Neural Architecture}
            \label{3layer network}
        \end{subfigure} 
        \\
        \begin{subfigure}[t]{1.0\linewidth}
            \includegraphics[width=\textwidth]{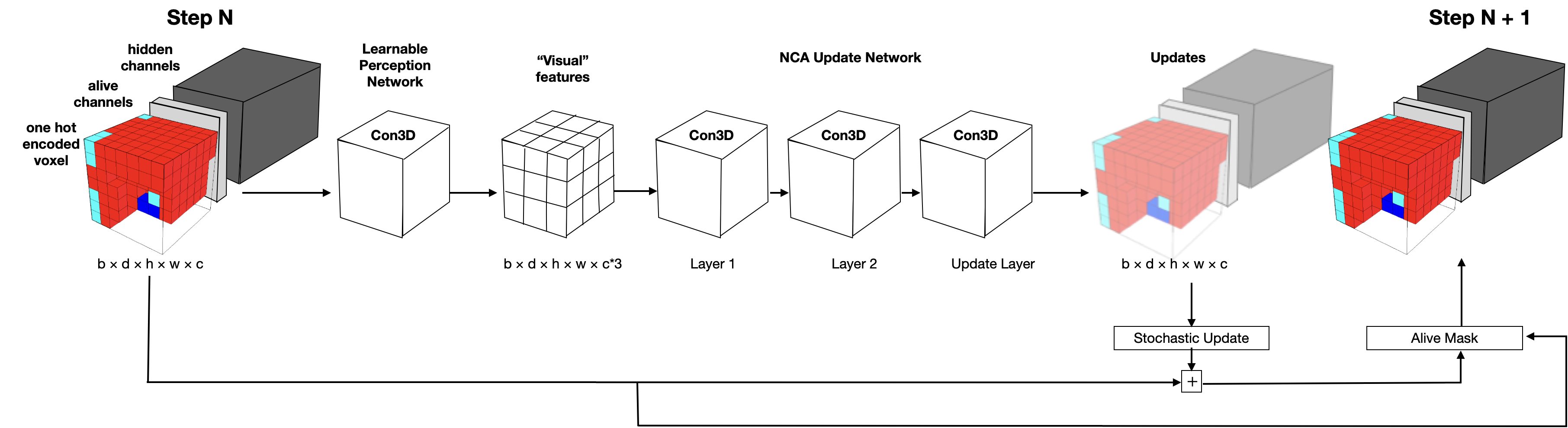}
            \caption{Differentiable Neural Architecture}
            \label{CNN network}
        \end{subfigure}
    \caption{{\bf Neural Cellular Automata Architecture.}   The evolutionary approach uses a simple three-layer network (a), while the differentiable programming approach is based on a more complex deep convolutional neural network (b).}
    \label{fig:network_architecture}
\end{figure}

\subsection{Neural Cellular Automata}

As previously discussed, Neural Cellular Automata (NCA) are Cellular Automata in which each cell state is represented by a vector of scalar values, and the update rules are parameterized by neural networks \citep{mordvintsev2020growing}. Our cell state vector consists of three parts: 1) The number of unique output types for each cell $k$, e.g. muscle, bone, etc., corresponds to the first $k$ channels, 2) the cell aliveness value and 3) a number of hidden states. The NCA grid is seeded with a single central cell at $t=0$ with its state vector set to all ones and all the other cell states to zero. The NCA update rule defines a recurrent additive function parameterized by a neural network,
\begin{align}
    h_t^i &= a_t^i \cdot (h_{t-1}^i + u_t^i \cdot f_\theta(\{h_{t-1}^j\}_{j \in N(i)})) \,,
\end{align}
where $h^i_t \in \mathbb{R^D}$ is the $D$ dimensional state of cell $i$ at time $t$, $f_\theta$ is the update rule, parameterized by a neural network with parameters $\theta$ and $N(i)$ is the set of indices of all the neighbors of cell $i$, including i. For a NCA defined on a 3D grid it corresponds to the cells in a $3 \times 3 \times 3$ neighborhood \citep{sudhakaran2021growing}. $a_t^i \in \{0,1\}$ and $u_t^i \in \{0,1\}$ are the alive and update masks respectively. 

The update mask is a random binary variable sampled with some probability $p$, so that $p(u_t^i = 1) = p$. The random update mask can be seen as a form of dropout \citep{srivastava2014dropout} which makes the NCA robust to noise and invariant to the exact order of the updates. In our experiments we use $p=0.5$.

The aliveness mask ensures that dead cells have an all zero state. A cell is considered alive if any of the cells in its $3 \times 3 \times 3$ neighborhood, including itself, have an aliveness value of more than 0.1. The aliveness value is the $k$'th entry of the cell state vector, so that $a_t^i = \max(\{h_{t-1}^j[k]\}_{j \in N(i)}) \geq 0.1$. The aliveness mask is efficiently computed using a 3D max pooling operation.

\section{Evolving Soft Robot Morphologies}
\label{sec3}

To confirm the promise of NCA for growing soft robots, we use a simple  three-layer networks with tanh activation functions (Fig.~\ref{fig:network_architecture}a). We experiment with both \textbf{feed forward} (the hidden layer is a linear layer) and \textbf{recurrent networks} (the hidden layer is an LSTM unit~\cite{Hochreiter1997}), which means that each cell has its own memory. Here, the dimension of the hidden layer is set to 64 unless otherwise noted. The recurrent setup is inspired by recent experimental reports that organisms store information about the original morphology in a distributed manner in the bioelectrical signaling networks \citep{Levin2017,McLaughlin2018}.

To evolve a NCA for initial robot growth, we use a simple genetic algorithm~\cite{Holland1992,Eiben2003} that has been shown to be effective in training deep neural networks~\cite{Such2017,risi2019deep}. The implemented GA variant performs truncation selection with the top $T$ individuals becoming the parents of the next generation. The following procedure is repeated at each generation: First, parents are selected uniformly at random. They are mutated by adding Gaussian noise to the weight vector of the neural network (its genotype): $\theta' = \theta + \sigma \epsilon$, where $\epsilon$ is drawn from normal distribution $N(0, I)$ and  $\sigma $ is set to $0.03$. Following a technique called elitism, top ${M}^{th}$ individuals are passed on to the next generation without mutation. 

\subsection{2D soft robots}
We first investigate the use of NCA in 2D. Here, robots have a maximum size of $7\times7$ voxels. Since the NCA used a Moore neighborhood, the input dimension of the neural network is $9\times2={18}$, which includes neighboring cell types and alpha values.  

\begin{figure}[htpb!]
    \centering
   
    \begin{minipage}{.5\linewidth}
        \begin{subfigure}[t]{.9\linewidth}
            \includegraphics[width=\textwidth]{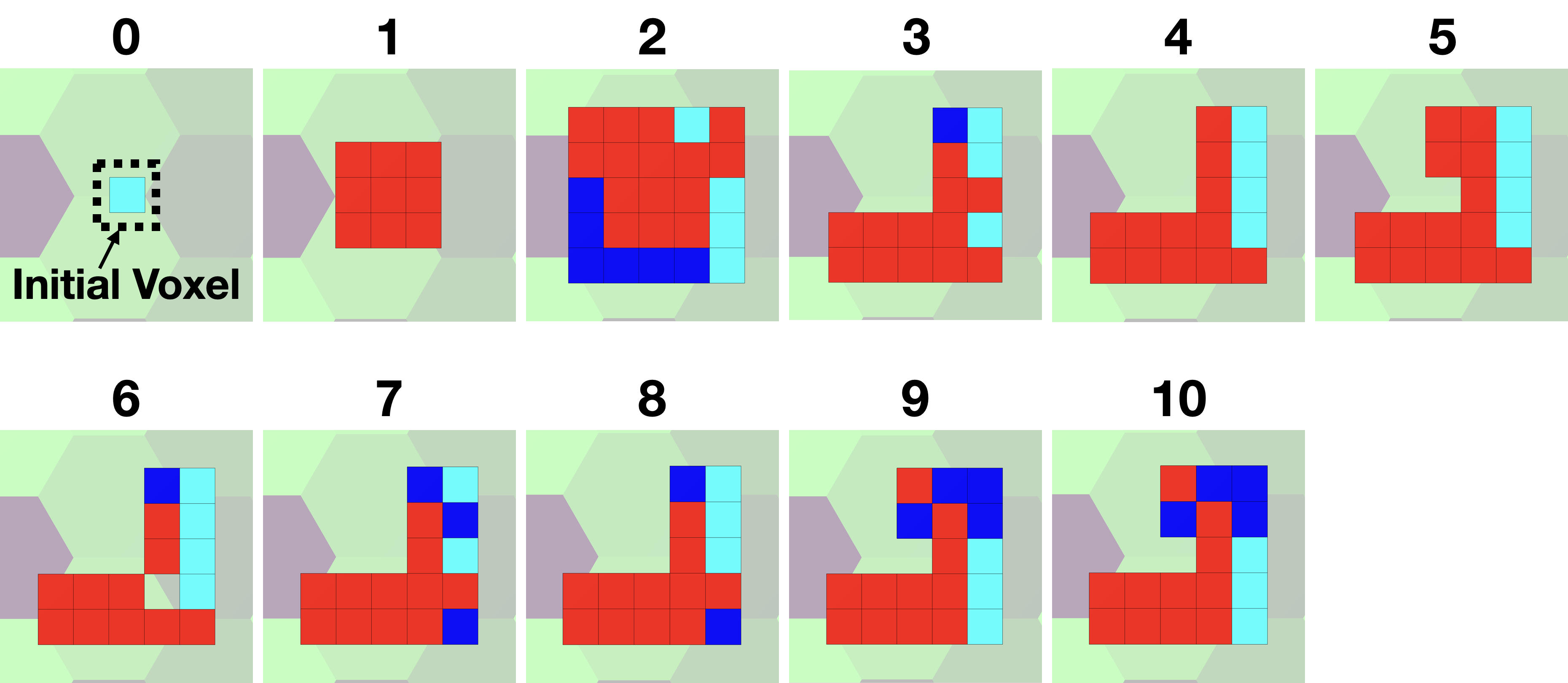}
            \caption{$2$D robot development}
            \label{2d creature development}
        \end{subfigure} \\
        \begin{subfigure}[b]{.9\linewidth}
            \includegraphics[width=\textwidth]{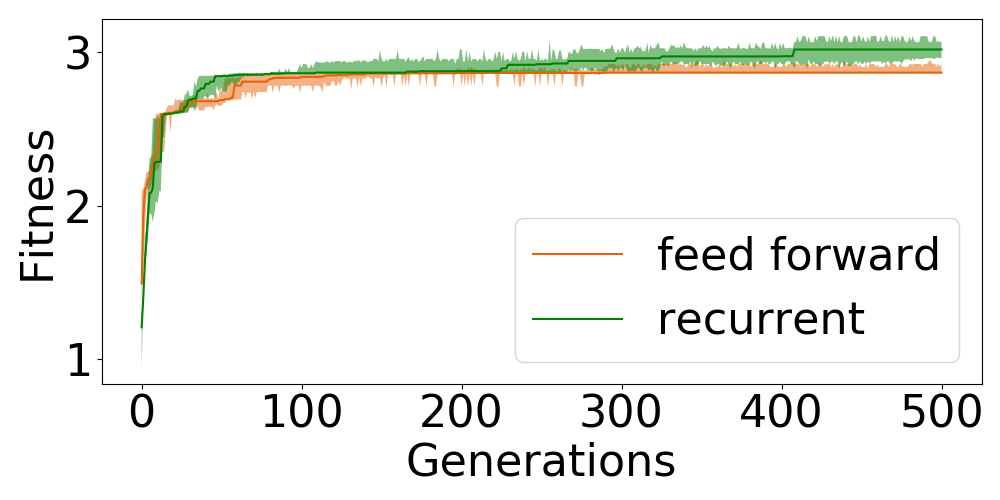}
             \caption{Training}
            \label{fitness2d}
        \end{subfigure} 
    \end{minipage}
     \begin{minipage}{.45\linewidth}
            \begin{subfigure}[t]{.9\linewidth}
                \includegraphics[width=\textwidth]{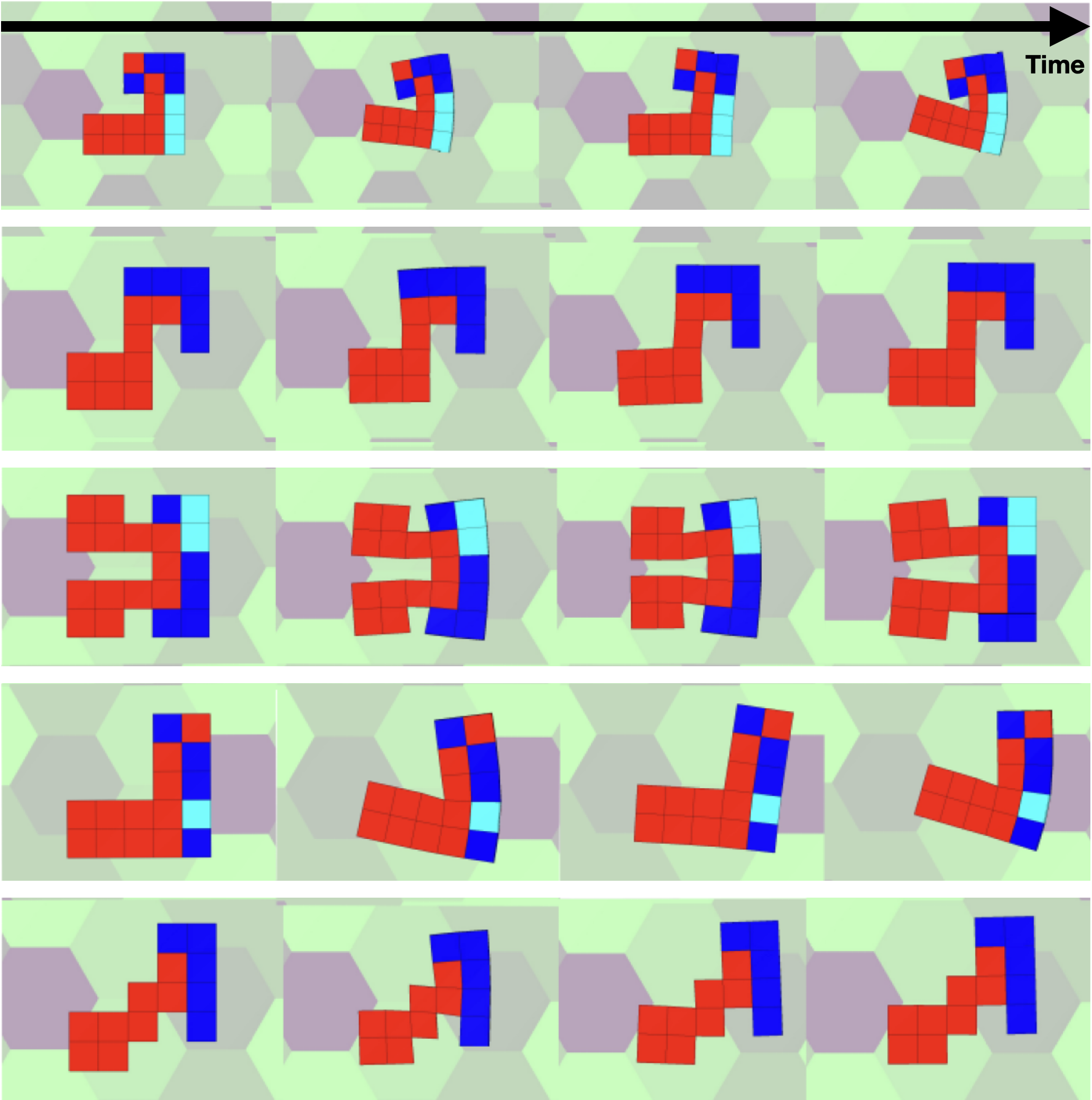}
                \caption{$2$D  robot locomotion}
                \label{2d creature locomotion}
            \end{subfigure}
    \end{minipage}
    \caption{{\bf Evolution of 2D soft robots} (a) Example of the development of 2D soft robots through a neural cellular automata. (b) Training fitness for the  recurrent/feed forward setup. (c) Time series of soft robot behaviors as they move from left to right. From top to bottom, we refer to them as Hook type, S-type, Biped, L-type, and Zigzag.}
    \label{2d results}
\end{figure}

The output layer of the NCA has a size of 6 neurons; 5 neurons for the different states of the cell (empty=0, light blue=1, dark blue=2, red=3, green=4) plus one alpha channel. The first single soft $\&$ passive cell (light blue) is placed at position $(3,3)$ and $10$ steps of development are performed. As result, 11 morphologies are obtained, see (Fig.~\ref{2d creature development}). Afterwards, the final grown robot is tested in the physical simulator (Voxelyze) and allowed to attempt locomotion for 0.25 seconds, i.e., 10 actuation cycles. The fitness of each robot is taken to be distance travelled by the robot from its starting point. These 2D experiments use a population size of $300$, running for 500 generations. One evolutionary run on 8 CPUs took around 12 hours.

Results were obtained from ten independent evolutionary runs, using both recurrent and feed forward networks. The training mean together with bootstrapped $95\%$ confidence intervals is shown in Fig.~\ref{fitness2d}. 
Evolution produced a variety of soft robots (Fig.~\ref{2d creature locomotion}). 
A  \qq{Hook} type is distinguished by its hook-like form and locomotion, which shakes the two sides of the hook and the proceeds to hook the remaining one side to the floor. 
The \qq{S} shaped-robot is distinguished by its sharp and peristaltic motion with amplitude in the same direction as the direction of travel.
The \qq{Biped} has two legs and its locomotion resembles that of a frog, with the two legs pushing the robot forward.  The \qq{L} type displays a sharp and winged movement. Finally, the \qq{Zig-zag} shows a spring-like movement by stretching and retracting the zigzag structure. Enabling the cells to keep a memory of recent developmental states through a recurrent network improved performance, although only slightly (Fig.~\ref{fitness2d}). Investigating what information the evolved LSTM-based network is keeping track of during development is an interesting future research direction. 

\subsection{3D soft robots}
\begin{figure}[t]
    \begin{subfigure}{.5\textwidth}
        \centering
        \includegraphics[width=1.0\linewidth]{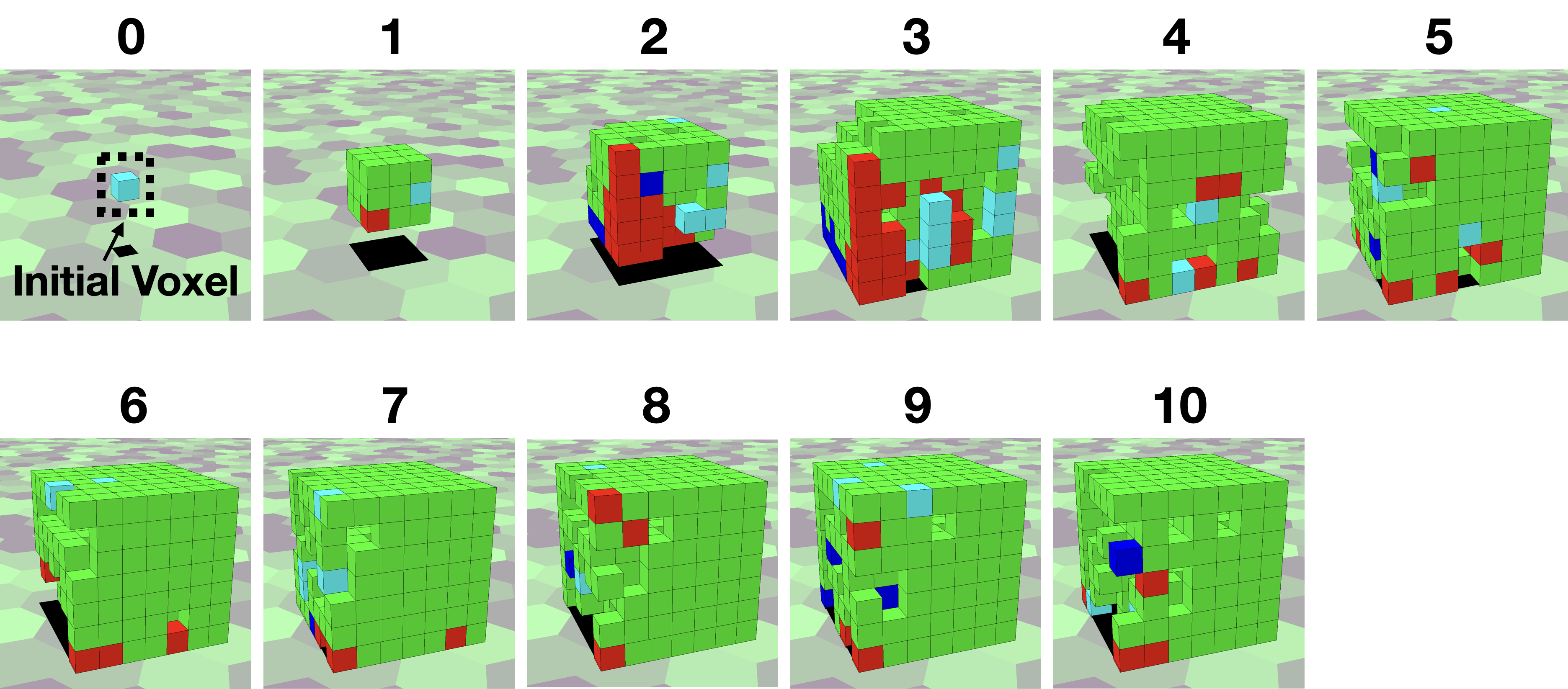}  
        \caption{3D soft robot  development }
        \label{3d creature development}
    \end{subfigure}
    \begin{subfigure}{.5\textwidth}
        \centering
        \includegraphics[width=1.0\linewidth]{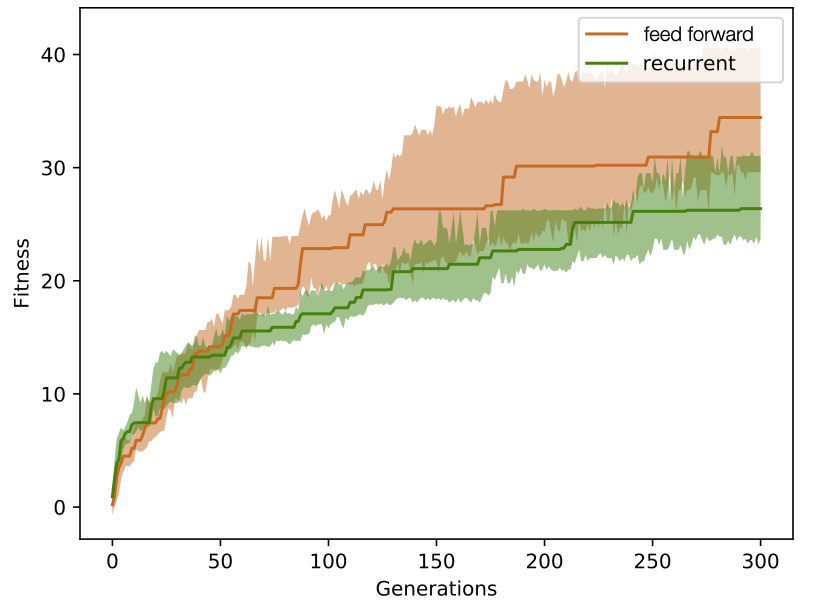}  
        \caption{Training}
        \label{fitness3d}
    \end{subfigure}
    \newline
    \begin{subfigure}{.5\textwidth}
        \centering
        \includegraphics[width=1.0\linewidth]{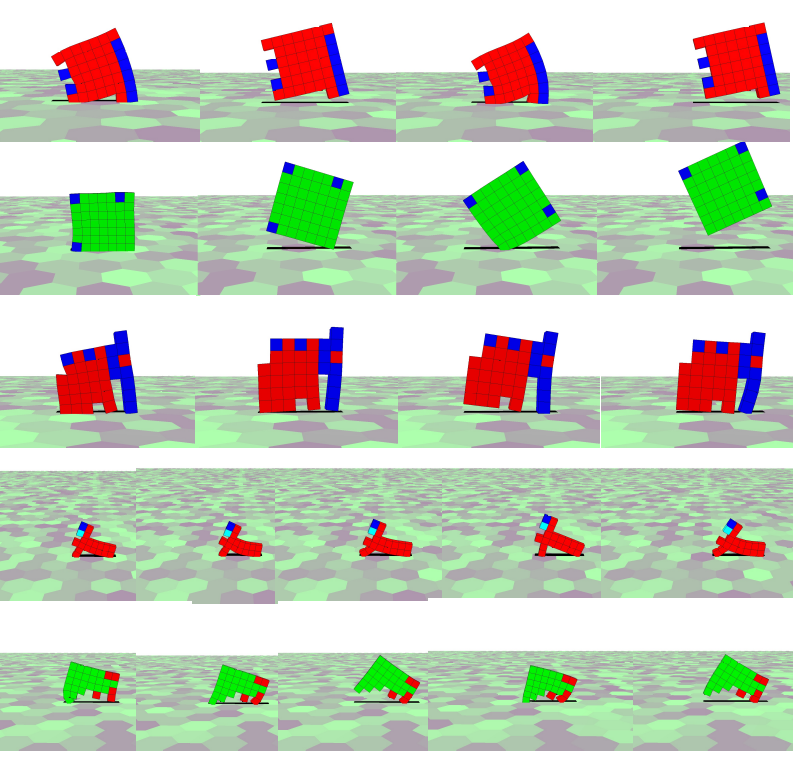}
        \caption{2D Group}
        \label{2D Phylum}
    \end{subfigure}
    \begin{subfigure}{.5\textwidth}
        \centering
        \includegraphics[width=1.0\linewidth]{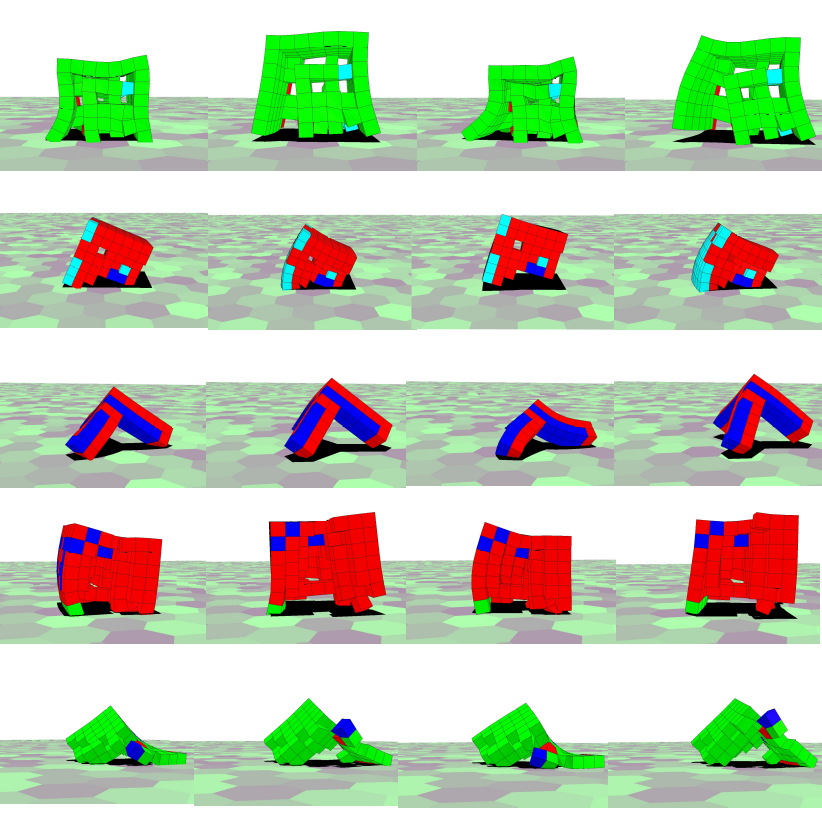}
        \caption{3D Group}
        \label{3D Phylum}   
    \end{subfigure}
    \caption{{\bf Evolution of soft robots} (a) A robots shown at different timesteps during its development. The initial light blue voxel is surrounded by a dotted line. (b) Fitness over generations for the  recurrent/feed forward setup. (c) Time series of common 2D soft robot behaviors as they move from left to right. From top to bottom, we refer to them as Jumper, Roller, Pull-Push, Slider, and Jitter. (d) Common grown 3D robots: Pull-Push, L-Walker, Jumper, Crawler, and Slider.}
    \label{3d results}
\end{figure}

In this section we now extend our methodology to grow 3D robots. For these 3D robots the maximum morphology size is  $9\times9\times9$. Because of the used Moore neighborhood, the input dimension of the neural network is $3\times9\times2=54$. The hidden layer is set to $64$. The output layer is set to $5+1=6$ dimensions with the number of states of the cell and the value of its own next step alpha value. 
The first single soft $\&$ passive cell is placed at position $(4,4,4)$ and $10$ steps of growth are performed 
(Fig.~\ref{3d creature development}). The final soft robot grown after $10$ steps is tested in Voxelyze and, as with the 2D robots, the distance of the robot's center of gravity from its starting point is used as part of the fitness function. Additionally, we include a voxel cost in the fitness calculation: $Fitness = (Distance) - (Voxels Cost)$. We added a ``voxel'' cost because preliminary results indicated that without this additional metric all the soft robots simply acquired a box-like morphology. Including the voxel cost metric increased diversity in the population. Note that voxel cost is the number of voxels that are neither empty nor dead.

For our 3D experiments, the evaluation time is increased to $0.5$s for $20$ actuation cycles to adjust for the increased complexity of the robots. Each generation has a population size of $100$ and the next generation is selected from the top $20\%$. The number of generations is set to 300. Note that both the generation number and population size are reduced from those values used in the 2D experiments as simulated the larger 3D robots has a higher computational cost. One evolutionary run on 1 CPU took around 80--90 hours.

Results are based on $24$ independent runs for both the  recurrent and feed forward treatment (Fig.~\ref{fitness3d}). 
Interestingly, the feed forward setup for the 3D robots has a higher fitness than the recurrent one, in contrast to the 2D soft robot results (Fig.~\ref{fitness2d}). We hypothesize that with the increased numbers of neighbors in 3D and more complex patterns, it might be harder to evolve an LSTM-based network that can use its memory component effectively. Because the dynamics of LSTM-based networks are inherently difficult to analyse, more experiments are needed to investigate this discrepancy further.

Similarly to CPPN-encoded soft robots \cite{cheney2014unshackling}, 3D robots grown by an evolved NCA (Figure~\ref{3d results}) can be classified into two groups: the first group is the two-dimensional group of organisms (Fig.~\ref{2D Phylum}), where planar morphology was acquired by evolution. Exemplary classes of locomotion in this group include the jumper, which is often composed of a single type of muscle voxel. Once a soft robot sinks down, it use this recoil to bounce up into the air and move forward. The morphology determines the angle of bounce and fall.
The Roller is similar to a square; it moves in one direction by rotating and jumping around the corners of the square. The Push-Pull is a widely seen locomotion style. A soft robot pushes itself forward with its hind legs. During this push, it pulls itself forward, usually by hooking its front legs on the ground.
The Slider has a front foot and a hind foot, and by opening and closing the two feet, it slides forward across the floor. The two legs are usually made of a single material. The Jitter moves by bouncing up and down from its hind legs to back. It has an elongated form and is often composed of a single type of muscle voxel.
The second group is the three-dimensional group of organisms, as shown in Fig.~\ref{3D Phylum}. The L-Walker resembles an L-shaped form; it moves by opening and closing the front and rear legs connected to its pivot point at the bend of the L.
The Crawler has multiple short legs and its legs move forward in concert.

\section{Training for Regeneration}

We now investigate the ability of the soft robots to regenerate their body parts to recover from morphological damage. We chose three morphologies from the previous experiments, which are able to locomote well and are as diverse as possible: the Biped (feed forward), Tripod (feed forward), and Multiped (recurrent). The morphologies of each of these three robots are shown in Fig.~\ref{regeneration set up} and the locomotion patterns in Fig.~\ref{3D Phylum}. We use two different approaches to train regeneration NCAs, the first one is an evolutionary approach as previously reported in \citep{horibe2021regenerating}. The second approach is based on gradient-based optimization, i.e.\ differentiable programming.  

\subsection{Evolutionary Training for Regeneration}
In these experiments, which have been first reported in our conference paper \cite{horibe2021regenerating},  we damage the morphologies such that one side of the robot was completely removed (Fig.~\ref{regeneration set up}). In the left side of these damaged morphologies, the cell states were set to empty and the maturation alpha values were set to zero. For the recurrent network, the memory of LSTM units in each cell were also reset to zero. 

\subsubsection{Network architecture and training method}
We initially attempted regeneration using the original NCAs of these three robots but regeneration failed and locomotion was not recovered. Therefore, we evolved another NCA, where the sole purpose was to regrow a damaged morphology. In other words, one NCA grows the initial morphology and the other NCA is activated once the robot is damaged. Fitness for this second NCA is determined by the voxel similarity between the original morphology and the recovered morphology (values in the range of $[0, 729]$). The maximum fitness of $ 9\times9\times9 = 729$ indicates that the regrown morphology is identical to the original morphology. We evolve these soft robots, which are allowed to grow for 10 steps, for $1,000$ generations with a population size of $1,000$. The next generation is again selected from the top $20\%$.

\begin{table}[h]
    \centering
    \begin{tabular}{ |p{3.2cm}||p{2cm}|p{1cm}|p{1.3cm}|p{1.3cm}|}
    \hline
    \multicolumn{2}{|c|}{}& \multicolumn{3}{|c|}{Locomotion}\\
    \hline
    Morphology (Network) & Similarity & Original & Damaged & Regrown\\
    \hline
    Biped (feed forward) & $98\%$ (718/729) & 40.4  & 27.2($67\%$) & 35.1($86\%$) \\
    Tripod (feed forward) & $99\%$ (728/729) &44.5 & 1.63($3.6\%$) & 20.3($45\%$) \\
    Multiped (recurrent) &$91\%$ (667/729)& 42.7 & 5.36($12\%$) & 9.6($22\%$) \\
    \hline
    \end{tabular}
    \caption{Morphology similarity and locomotion recovery rate.}
    \label{regeneration locomotion}
\end{table}

\subsubsection{Morphology similarity and locomotion recovery}
\begin{figure}[htpb!]
    \begin{subfigure}{.5\textwidth}
        \centering
        \includegraphics[width=.8\linewidth]{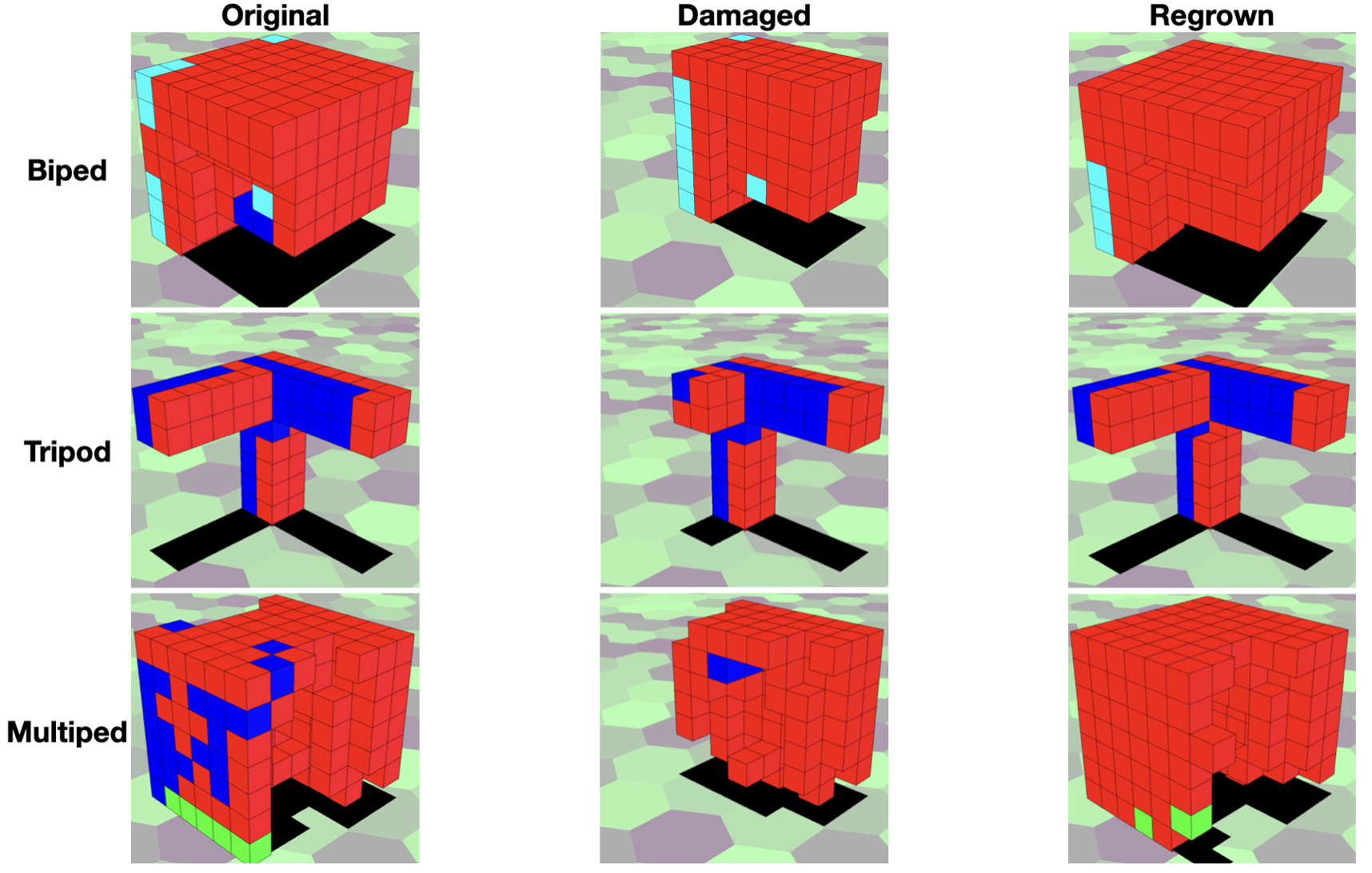}
        \caption{Original, damaged and regrown}
        \label{regeneration set up}
    \end{subfigure}
      \begin{subfigure}{.5\textwidth}
        \centering
        \includegraphics[width=1\linewidth]{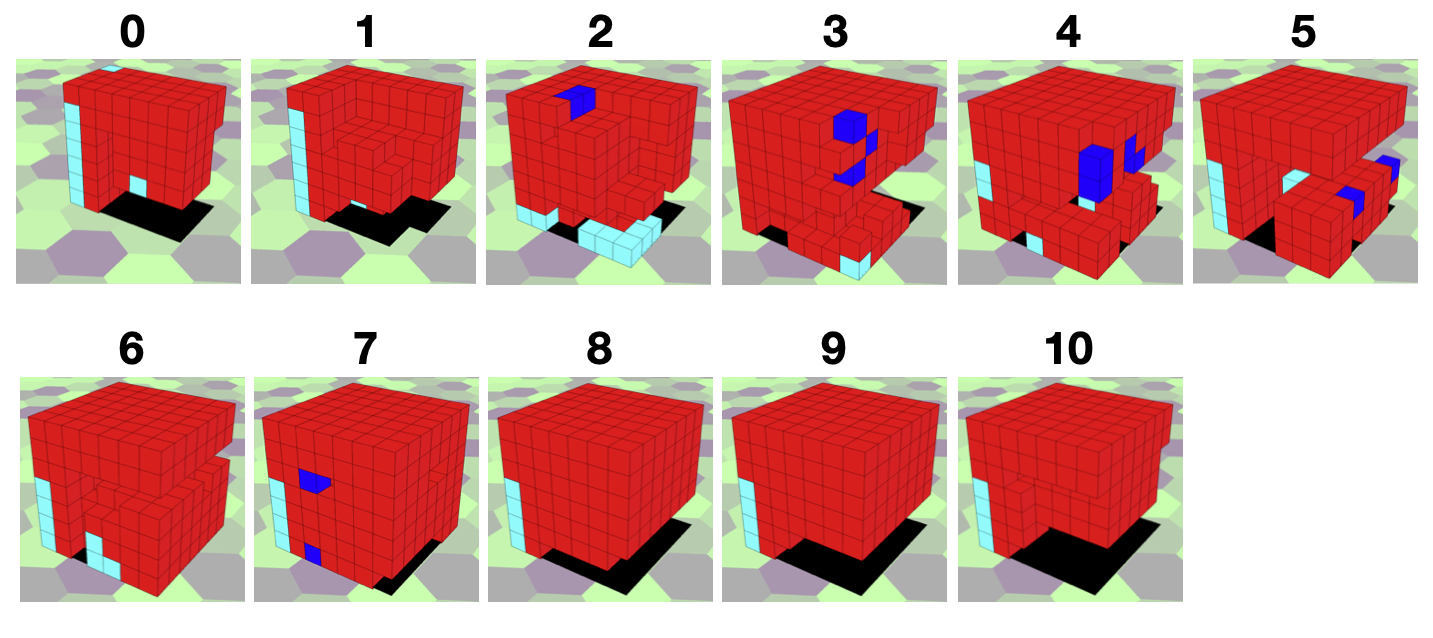}
        \caption{Biped regereration}
        \label{biped regeneration}
    \end{subfigure}
    \newline
    \begin{subfigure}{.5\textwidth}
        \centering
        \includegraphics[width=1\linewidth]{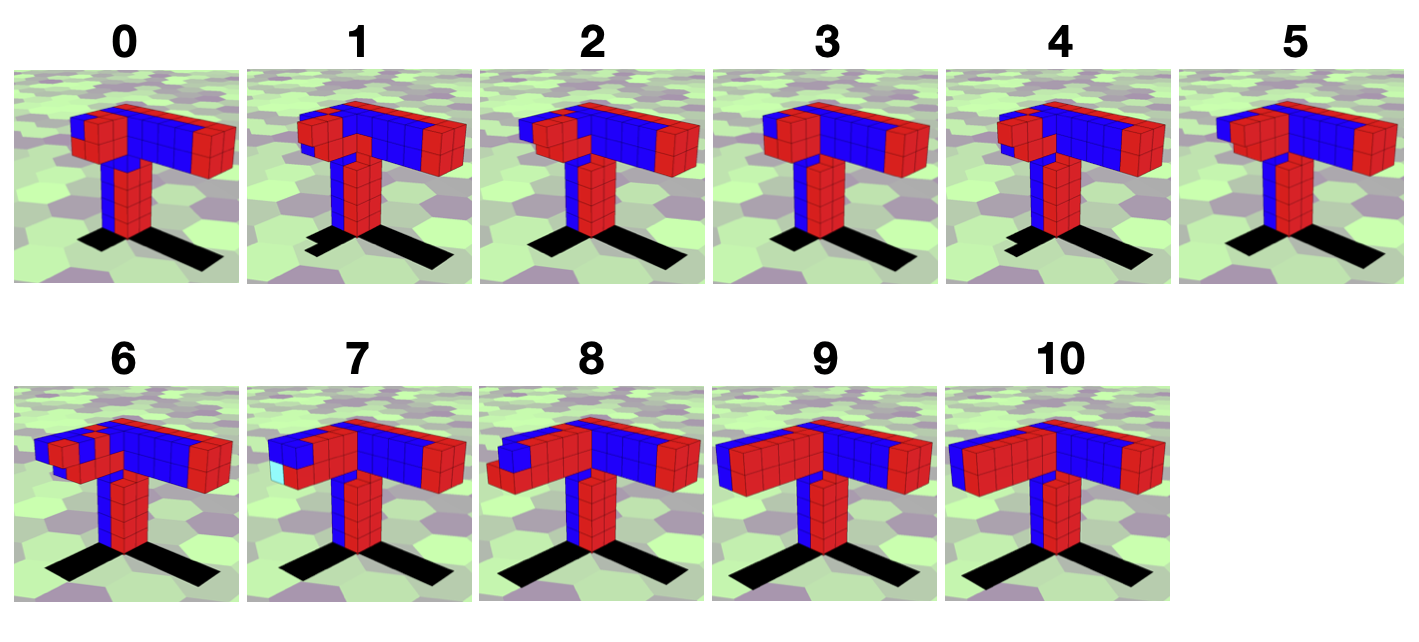}
        \caption{Tripod regereration}
        \label{Tripod regeneration}
    \end{subfigure}
    \begin{subfigure}{.5\textwidth}
        \centering
        \includegraphics[width=1\linewidth]{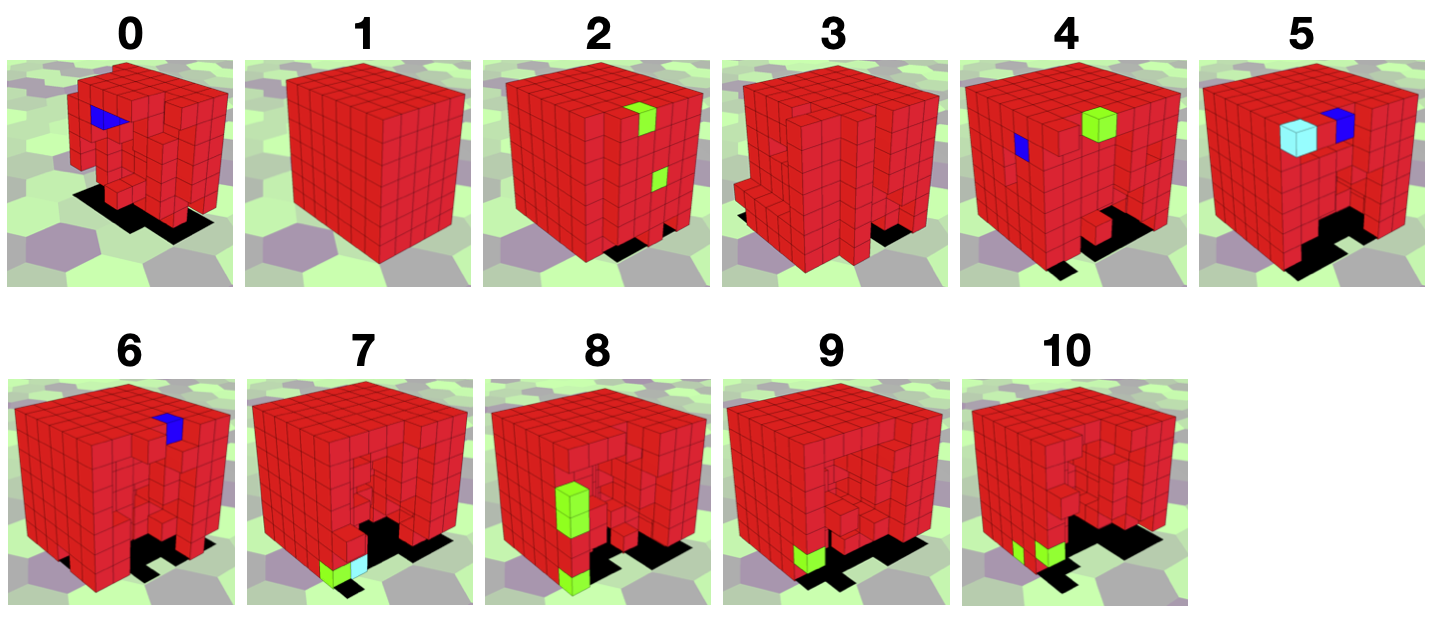}
        \caption{Multiped regereration}
        \label{Multiped regeneration}
    \end{subfigure}
    \newline
    \begin{subfigure}{1\textwidth}
        \centering
        \includegraphics[width=1\linewidth]{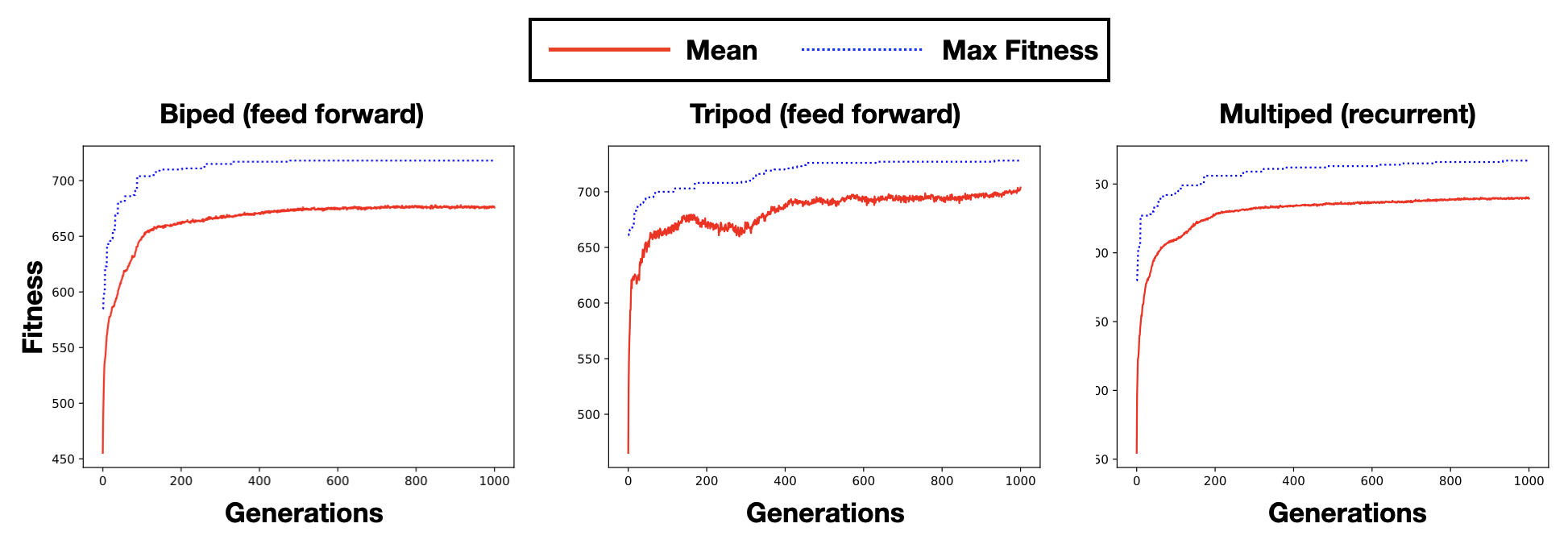}
        \caption{Fitness function of each morphology}
        \label{regeneration fitness}
    \end{subfigure}
    \caption{{\bf Soft robots evolved for regeneration} (a) Original, damaged, and regrown morphology. (b)-(d) Soft robot development after damage shown at different timesteps. (e) Training performance for recurrent/feed forward setup.}
    \label{regeneratio}
\end{figure}

For all three morphologies, we trained both feed forward and recurrent  NCAs. The best performing network types for damage recovery were consistent with the original network type for locomotion in all morphologies (biped = feed forward, tripod = feed forward, multiped = recurrent ). 
Training for the Biped (feed forward), Tripod (feed forward) and Multiped (recurrent) takes 45, 23, and 100 hours, respectively, using a single CPU core (2.7 GHz Intel Xeon E5).
The results with the highest performing network type are summarised in Table~\ref{regeneration locomotion} and damaged morphologies for each of the robots are shown in Fig.~\ref{regeneration set up}. 
The results indicate that the Multiped was the hardest to reproduce, followed by the Biped and then the Tripod. The Tripod had a higher similarity than the other morphologies and the NCA almost completely reproduced the original morphology with the exception of one cell. We hypothesise that regeneration for the Tripod is easier because it only requires the regrowth of one leg, a simple rod-like shape with only a few cells. 

For comparison, we then measured the locomotion of the original, damaged, and regrown morphology with an evaluation time of $0.5$s for $10$ cycles in Voxelyze. The ratio of regrowth and travel distance to the original morphology are shown in Table~\ref{regeneration locomotion} and its locomotion in Fig.~\ref{fig:all locomotion}. The damaged Biped maintained 67$\%$ of its original locomotion ability;  it replicated a similar locomotion pattern to the one observed in the L-Walker. As the Tripod lost one of its three legs, it was incapable of successful locomotion. Furthermore, the Multiped lost all locomotion -- the robot simply collapsed at the starting position. 

These results suggest that the location of the damage is important in determining how much the robot loses in terms of locomotion performance. For instance, in the case of the Biped, the left hand side and right hand side are symmetrical. This means that when the left hand side was removed, the right hand side was able to locomote in the same, almost unaffected way. Therefore, despite having the lowest similarity value  between the initial and regrown morphologies, there is little loss in performance. 
In contrast, the Tripod regained less than half the locomotion of the original morphology, despite regaining its original morphology almost completely. It would appear that the one voxel it is unable to regenerate is necessary to prevent the robot from spinning and thus from moving forward.

Using this method of training two neural networks via an evolutionary approach shows potential for soft robotic regeneration with NCAs. However, only one type of damage scenario is investigated, and even in this case our method is not successful at recovering the functionality of the robot. Therefore, in the next section, we investigate a new method of training one neural network for growth \emph{and} recovery through gradient-based optimization. 

\subsection{Regeneration through differentiable programming}
In the gradient-based training, \emph{one} NCA is trained to grow a particular target structure (one of the three evolved robot morphologies) while being resilient to damage. 

\subsubsection{Network architecture and training method}
Here we use a similar three-dimensional convolutional network than the one introduced by Sudhakaran et al.~ \citep{sudhakaran2021growing}. This network (Fig.~\ref{fig:network_architecture}b) is an extension of the 3D network proposed by Mordvintsev et al.~\citep{mordvintsev2020growing}, which has been shown to exhibit robust regeneration 2D patterns against various types of damage. The input of the NCA first passes through a perception net implemented with a 3D convolutional layer with {\fontfamily{qcr}\selectfont kernel size=3, stride = 1} and  {\fontfamily{qcr}\selectfont output channels = cell state the NCA * 3}. Next, its output passes through three linear 3D convolutional layers with {\fontfamily{qcr}\selectfont kernel size=1, stride = 1} of the NCA update network. Following \cite{mordvintsev2020growing,sudhakaran2021growing}, we apply stochastic updates, i.e.\ per-cell dropout. 
An ``alive mask'' is implemented by multiplying the MaxPool layer of the update by a Boolean filter, which is 1 if the value of the living channel is $ < $ 0.1, and 0 otherwise. All network weights are   initialization with standard normal = 0.1 and mean = 0.0. 


The additive update rule is inspired by residual networks \citep{he2016deep}, and helps with stabilizing the gradient during training. The update rule can be efficiently implemented using a single 3D convolution with a filter size of $3 \times 3 \times 3$ followed by a non-linearity and any number of 3D convolutions with $1 \times 1 \times 1$ filter sizes, followed by non-linearities. A final linear $1 \times 1 \times 1$ convolution maps back to the NCA hidden state size. In our experiments we follow \citep{sudhakaran2021growing} and use two $1 \times 1 \times 1$ convolutions with 64 channels each, and ReLU non-linearities \citep{nair2010rectified}.

During training the NCA is run for a random number of steps, $T$ between 48 and 64, after which the loss is calculated. We use the loss from \citep{sudhakaran2021growing}, which contains two parts,
\begin{align}
    \mathcal{L} &= \sum_i \frac{1}{2} \text{CE}(y_i, t_i) + \text{IOU}(y_i, t_i)
    \label{eq:loss}
\end{align}
where $y_i = \text{softmax}(h_T^i[:k])$ is the softmax of the first $k$ channels of the final cell state, $\text{CE}$ is the categorical cross entropy loss, where $t_i \in \{1, ..., k\}$ are the cell type targets. In practice we split the cross entropy loss of empty voxels and the cross entropy loss of the rest of voxels. This split helps with training stability, as the loss for non-empty voxels does not overshadow the loss for the empty voxels. $\text{IOU}$ is the Intersection Over Union loss which measures the absolute distance between non empty voxels and empty voxels. Given the loss the gradients are computed with back-propagation using automatic differentiation and the NCA parameters are trained with a gradient-based approach, using the Adam optimizer \citep{kingma2014adam}.

In this case the NCA  quickly learns to grow the target structure, but is not stable when run for more time steps than it was trained for and can only recover from limited damage. In order to learn a NCA that is stable over thousands of time steps and capable of regenerating from severe damage we modify the training, following the approach outlined in \citep{mordvintsev2020growing}.

\begin{algorithm}
\caption{NCA pool training with damage} \label{alg:nca_training}
\begin{algorithmic}
\State $pool \gets init()$
\While{not converged}
    \State $idx, seeds \gets \text{sample}(pool, 5)$
    \State $losses \gets \mathcal{L}(seeds)$
    \State $seeds \gets \text{sort}(seeds, losses)$ \Comment{Sort seeds according to loss descending}
    \State $seeds[0] = initial$ \Comment{Replace worst seed with initial seed}
    \State $seeds[-2:] = \text{damage}(seeds[-2:])$ \Comment{Apply damage to two best seeds}
    \State $T \sim \mathcal{U}[48, 64]$
    \State $seeds \gets \text{NCA}(seeds, T)$ \Comment{Run the NCA for $T$ steps}
    \State $loss \gets \sum \mathcal{L}(seeds)$
    \State $\text{optimize}(\text{NCA}, loss)$ \Comment{Train the NCA with gradient descent}
    \State $pool[idx] \gets seeds$ \Comment{Replace samples from pool with new seeds}
\EndWhile
\end{algorithmic}
\end{algorithm}

Instead of always training from the same seed, at each training step we sample a batch of 5 random seeds from a pool of size 32. Of the 5 sampled seeds, we replace the one with the highest loss with a single cell seed, and apply damage to the two seeds with the lowest losses. We damage the seed by setting all the cell states to zero in a sphere with radius three, except the first channel, which corresponds to the empty block, which we set to one. After the training step we replace the sampled seeds in the pool with the output of the NCA at the last step, $h_t$. The pool is initialized with 32 single cell seeds. See Algorithm \ref{alg:nca_training} for details.



\subsubsection{Morphology similarity and locomotion recovery}
\begin{figure}[htpb!]
    \centering
    \includegraphics[width=1\linewidth]{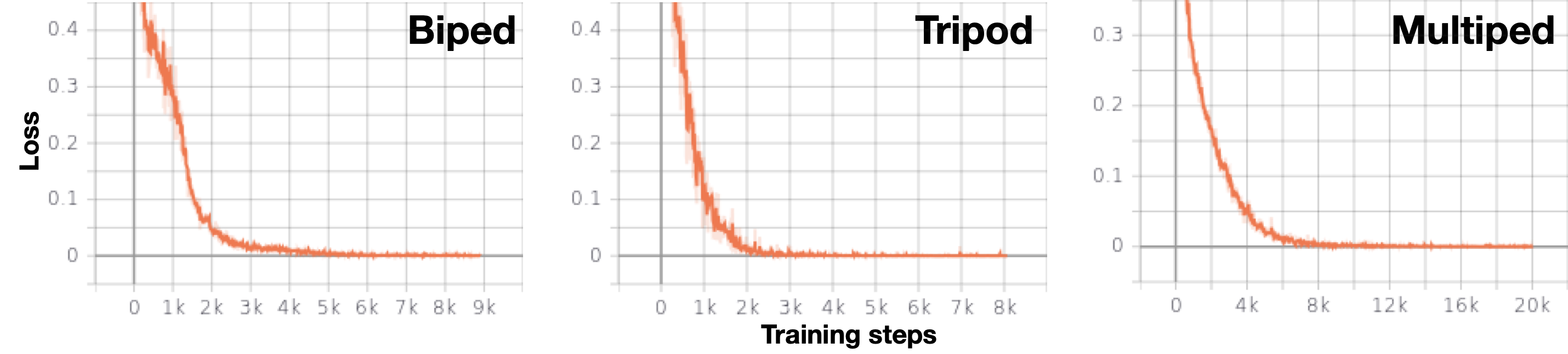}
    \caption{{\bf Training curves for gradient-based optimization of 3D NCAs.}  From left to right: Biped, Tripod, and Multiped. The loss decreases rapidly for all three morphologies.}
    \label{fig:dl_training}
\end{figure}

We performed 20,000 steps of learning unless the loss function reaches 0. Training for 8903, 8066, and 20000 steps for Biped, Tripod, and Multiped takes 15, 12, and 48 minutes using a GeForce GTX 1650 (TU117), respectively. The training curves for each morphology are shown in Fig. \ref{fig:dl_training}, displaying a rapid decrease in the loss.

\begin{figure}[htpb!]
    \centering
    \includegraphics[width=1\linewidth]{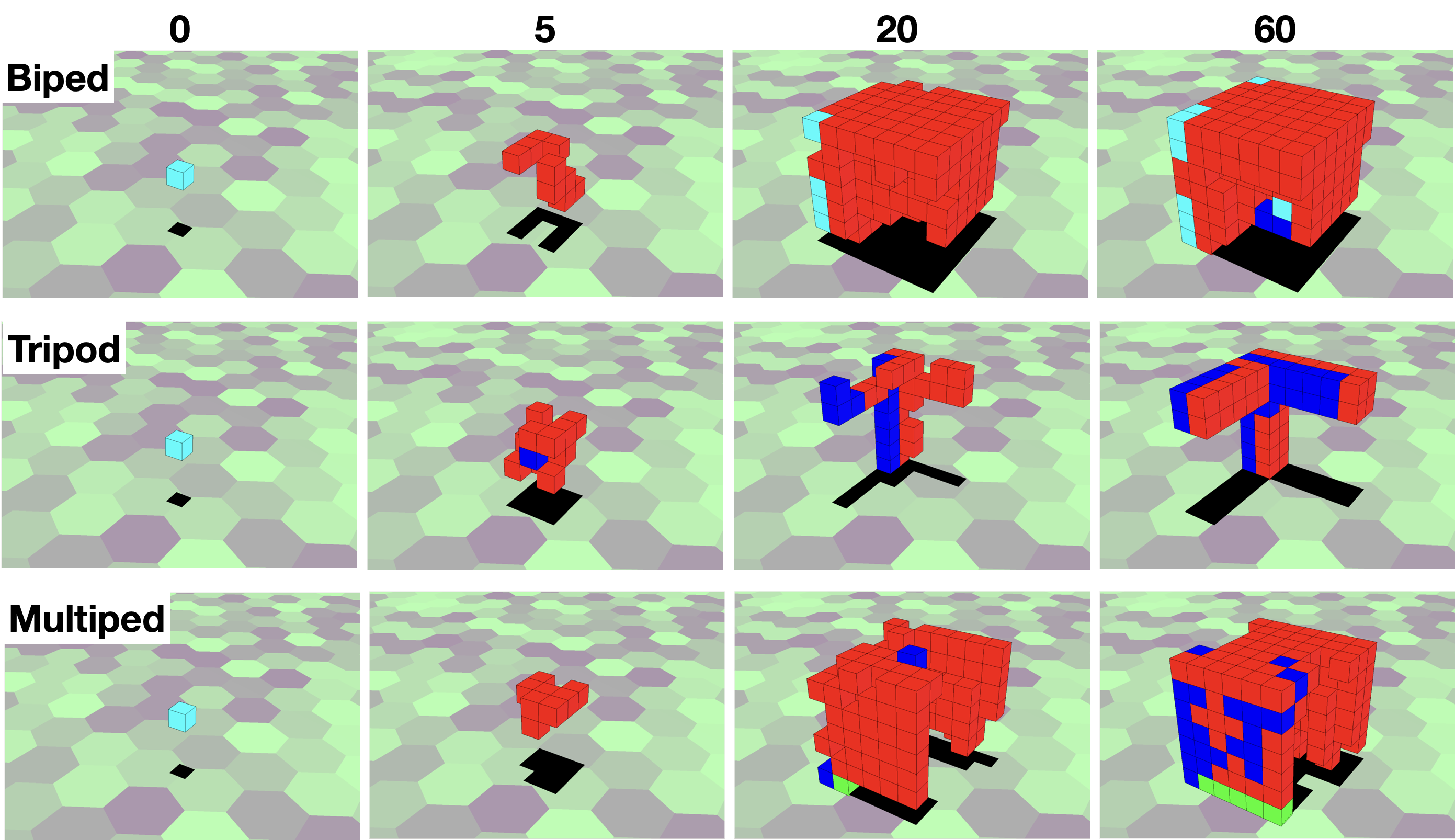}
    \caption{{\bf Growing soft robots from a single cell with gradient-based trained NCAs.}}
    \label{fig:gl_growth}
\end{figure}

We first examined whether the trained network is able to grow the original target morphology from a single cell. The first single soft $\&$ passive cell is placed at position $(3,3,3)$  with alive channel = 1 and 10 hidden states with random numbers 0 to 1. All other cells are empty voxels with alive channel and 10 hidden states set to 0. After running the NCA for 60 steps from the initial condition, we confirmed all the grown  morphologies completely matched the original morphologies (Fig.~\ref{fig:gl_growth}).

\begin{figure}
    \centering
    \includegraphics[width=1\linewidth]{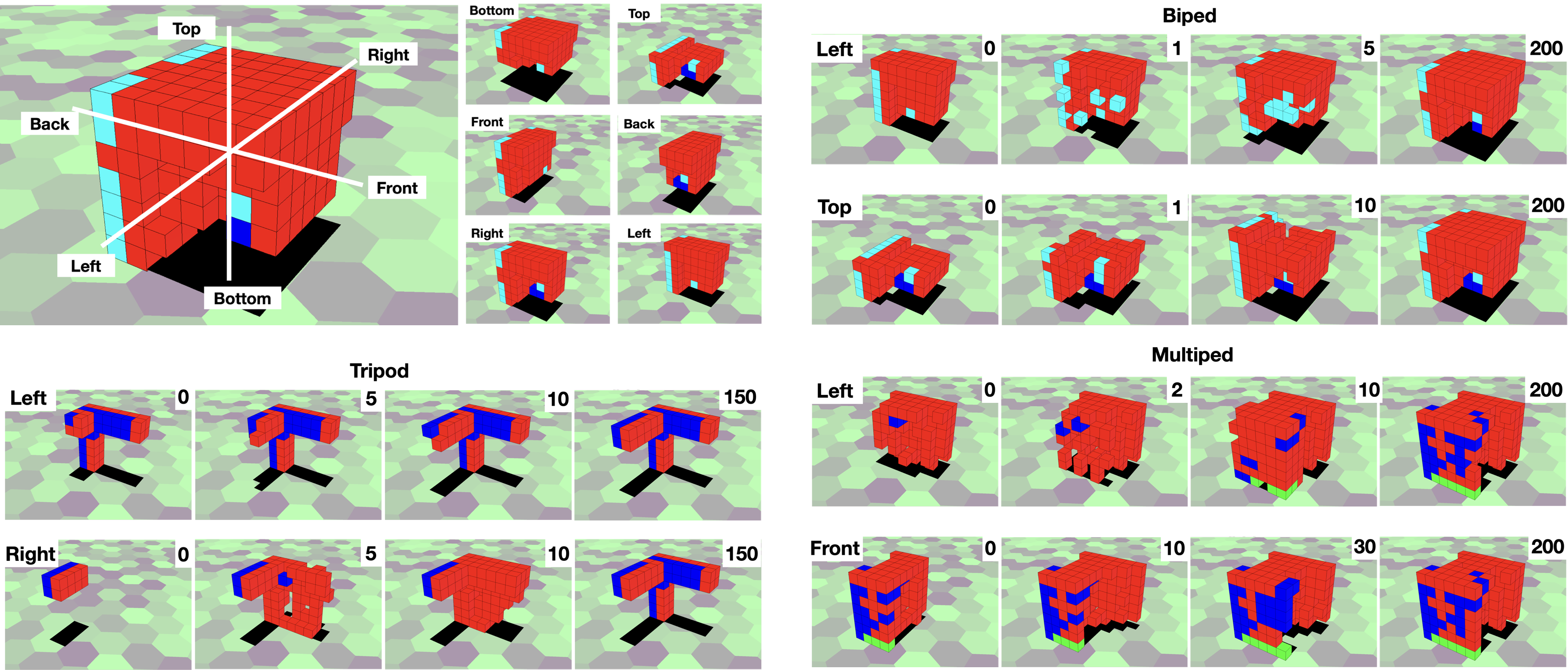}
    \caption{{\bf Regeneration against various damage with NCAs trained by gradient-based optmization.}  Definition of damage types are shown in the upper left panel. The other panels show the regeneration during different timesteps for different types of damage for  the Tripod (left \& right damage), Biped (left \& top damge), and Multiped (left \& front damage).} \label{fig:morphological resilienc}
\end{figure}

\begin{table}[]
    \centering
\begin{tabular}{ |p{2.4cm}||p{1cm}|p{1cm}|p{1cm}|p{1cm}|p{1cm}|p{1cm}|}
 \hline
 Morphology(State) & Left & Right & Top  & Bottom  & Front  & Back \\
 \hline
 Biped(Damage)  & $67\%$ & $17\%$ & $9\%$ & $1\%$ & $23\%$& $1\%$\\
 Biped(Regrown)  & $98\%$ & $98\%$ & $100\%$& $103\%$& $94\%$ & $93\%$\\
 \hline
 Tripod(Damage) & $5\%$& $14\%$ & $22\%$& $22\%$ & $16\%$& $12\%$\\
 Tripod(Regrown) & $100\%$& $100\%$ & $100\%$ & $100\%$ & $100\%$& $100\%$\\
 \hline
 Multiped(Damage)& $12\%$& $3\%$ & $10\%$ & $3\%$& $15\%$ &$10\%$\\
 Multiped(Regrown)& $98\%$ & $43\%$ & $25\%$ & $95\%$& $83\%$ &$97\%$\\
 \hline
\end{tabular}
    \caption{Recovery of locomotion ability of the regrown morphologies after various types of damage (left, right, top, bottom, front, back).}
    \label{tab:locomotion}
\end{table}

Next, we evaluated the system's ability for morphological regeneration and locomotion recovery. 
Based on the three spatial axes defined in Fig. 2, Left-Right, Top-Bottom, and Back-Front, six types of cuts were made to each of the three soft robots. The cell types were set to empty and  the active channel and hidden states to 0 for three of the seven columns of each axis. Starting from one of the damaged morphologies, the NCA was run for 200 steps for the Biped / Multiped and 150 steps  for the Tripod to see if it can regrow the original morphology (Fig.~\ref{fig:morphological resilienc}). 
\begin{table}[htpb!]
    \centering
\begin{tabular}{ |p{3cm}||p{0.8cm}|p{0.8cm}|p{0.8cm}|p{0.8cm}|p{0.8cm}|p{0.8cm}|}
 \hline
  Morphology(Network) & Left & Right & Top  & Bottom  & Front  & Back \\
 \hline
 Biped(CNN)  & $99.4\%$ & $ 99.7\%$  &  $100\%$ & $99.1 \%$  & $99.4\%$ & $99.1\%$ \\
 Tripod (CNN) & $100\%$ &  $100\%$  &  $100 \%$ &  $100\%$  & $100 \%$ &$100\%$ \\
 Multiped (CNN)&  $99.7\%$ &  $99.1\%$ & $97.9 \%$& $98.2 \%$&  $97.9\%$ & $98.8\%$\\
 \hline
\end{tabular}
    \caption{Similarities of the morphologies that were regrown after different types of damage (e.g. left, right, top, bottom, front, back) to the  original morphologies.}
    \label{tab:Similarity}
\end{table}

The similarities of the regrown morphologies to the  original morphologies are shown in Table~\ref{tab:Similarity}. The Tripod was easier to regenerate than the other two morphologies and completely regrew the original morphology for all types of damage. During regrowth, the morphology quickly matched (within 10 steps) the original one almost perfectly, but required 150 developmental steps for a perfect match. The Biped only achieved 100\% recovery for the top damage. However, it did not improve the similarity for the other types of damage even when run for more than 200 steps. The Mutiped did not achieve 100\% regeneration for any damage.

We measured the locomotion of original, damaged, and regrown morphology with an evaluation time of 0.5s for 10 cycles in VoxCad. The travel distance for original morphology is shown in Table~\ref{regeneration locomotion}, and the rate of locomotion for damaged and regrown morphology is shown in Table~\ref{tab:locomotion}. Biped's left damage remained at 67\% as in the evolutionary calculation (Table~\ref{regeneration locomotion}), while other damage critically lost its locomotion ability. For the Tripod, all damaged morphologies fatally lost their movement ability but regained it 100\%, completely regrowing their original morphology after all types of damage. In Multiped, the soft robot regained more than 80\% of its mobility, except for Right and Top damage, where the soft robot collapsed in the middle due to the placement of a small number of cells and could not move any further, resulting in a small mobility value.

\subsubsection{Comparing regeneration through evolutionary vs.  differentiable programming} 

The similarities of the regrown morphologies to the original morphologies after Left damage were higher for all three morphologies for the differentiable training compared to the evolutionary training (Biped  99.4\% compared to 98\%, Tripod 100\% compared to 99\%, and Multiped 99.7\% compared to 91\%).  The results are summarised in Table~\ref{regeneration locomotion} and \ref{tab:Similarity}.  The NCAs produced by the differentiable training show overall high robustness against all six damage types,  reaching a similarity of more than  95\% for all three morphologies  (Table~\ref{tab:Similarity}).

\begin{figure}
    \centering
    \includegraphics[width=1\linewidth]{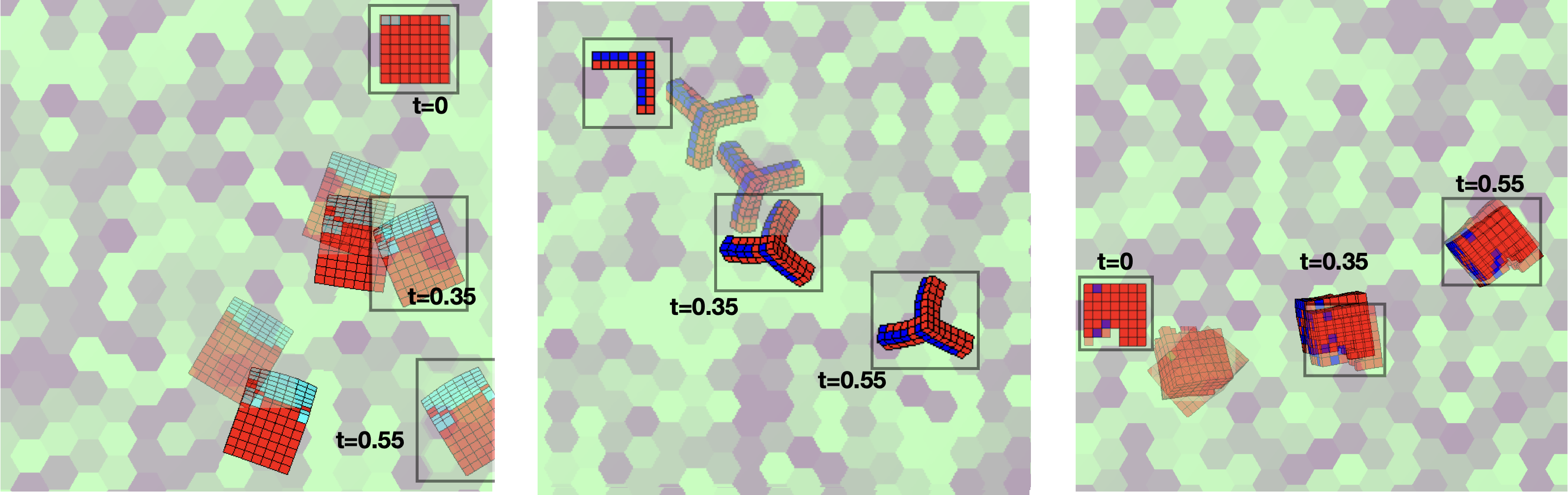}
    \caption{{\bf Recovery of locomotion using evolutionary algorithm and differentiable programming.} From left to right: Biped, Tripod, and Multipod. The regrown morphologies by evolutionary training are shown  transparent, while the regrown morphologies by differentiable programming are shown transparent with a frame around them.}
    \label{fig:all locomotion}
\end{figure}

Fig.~\ref{fig:all locomotion} shows the locomotion of the  original morphology, regrown morphology by evolutionary training, and regrown morphology by differentiable programming under the left damage condition. For all three soft robots, the regrown morphologies obtained through differentiable programming travelled further than that obtained by evolutionary optimization. 

It is important to note that it is difficult to fairly compare these two different optimization approaches. Both methods not only use different optimization algorithms but also different neural network architectures and hyperparameters. However, while the efficiency of our genetic algorithm can likely be improved, the results reported here mirror similar results from previous research: artificial evolution often struggles when tasked to perform supervised learning \cite{woolley2011deleterious} and even learning to grow simple 2D structures can be challenging \cite{nichele2017neat}. In cases where a target structure is given, supervised learning through gradient-based optimization has shown to be the efficient method of choice, and has allowed significantly more complex 2D \cite{mordvintsev2020growing} and 3D patterns \cite{sudhakaran2021growing} to be grown through neural networks than have previously been possible. On the other hand, evolutionary algorithms have shown to be better suited for open-ended search \cite{pugh2016quality} (instead of training towards a target) and the creative discovery of artefacts and behaviours  \cite{lehman2020surprising}.  The approach presented in this paper thus combines the advantages of both methods and we hope will spur the development of more of such hybrid methods.

\section{Discussion and Future Work}
The growth and regeneration abilities of complex multicellular tissues are specific characteristics of biological systems, allowing flexible adaptation to their environments. 
In this paper, we developed a new method for simulated soft robots to regenerate from damage based on neural cellular automata. The approach first evolves various growing soft robots and in a second steps trains them to being able to regenerate the original morphology after different types of damage through differentiable programming. Although complete regrowth to the original morphology was not always achieved,  all regrown soft robots regained more than 80\% of their travel distance after all types of damage, except for Right and Top damage to the  Multiped. These results suggest that growth can increase the evolutionary diversity of soft robot morphologies and that regeneration can provide resilience to damage.

While the locomotion and regeneration tasks in this paper are relatively simple, they open up exciting future  directions such as object manipulation, adaptation to environmental changes, task-based transformation, and self-replication.

Recently, soft robots designed using computer simulations have been recreated in real robots using a variety of materials~\cite{Howison2020}. With  the development of material science, a variety of soft robots that can change their shape have also been created~\cite{El-Atab2020} 
and hybrid robots with dynamic plasticity are being fabricated \cite{kriegman2020scalable}. In the future, it may be possible to create hybrid robots with living tissue that can grow spontaneously and recover functionality after severe damage, by building on the proposed approach. Because the method presented in this paper only relies on the local communication of cells, it could be a promising avenue to explore for the next generation of these hybrid robots.

\section{Acknowledgements}
This work was supported by the Tobitate! (Leap for Tomorrow) Young Ambassador Program, a DFF-Research Project1 grant (9131-
00042B), and KH's Academist supporters\footnote[1]{https://academist-cf.com/projects/119?lang=en}(Takaaki Aoki, Hirohito M. Kondo, Takeshi Oura, Yusuke Kajimoto,Ryuta Aoki).

\begin{appendices}

\end{appendices}


\bibliography{sn-bibliography}


\end{document}